\documentclass{article} 
\usepackage{iclr2023_conference_tinypaper,times}

\usepackage{amsmath,amsfonts,bm}

\def\eqref#1{equation~\ref{#1}}

\def\1{\bm{1}}

\DeclareMathAlphabet{\mathsfit}{\encodingdefault}{\sfdefault}{m}{sl}
\SetMathAlphabet{\mathsfit}{bold}{\encodingdefault}{\sfdefault}{bx}{n}

\DeclareMathOperator*{\argmin}{arg\,min}

\usepackage{multirow}
\usepackage{hyperref}
\usepackage{url}
\usepackage{graphicx}
\usepackage{subcaption}
\usepackage{algorithmic}

\usepackage[linesnumbered,lined,boxed,commentsnumbered]{algorithm2e}
\title{Key Patch Proposer: Key Patches Contain Rich Information}

\author{Jing Xu $^1$\thanks{ This work is done during Jing Xu's internship at Institute for AI Industry Research, Tsinghua University. } , Beiwen Tian$^2$ \& Hao Zhao$^2$ \\
Nanjing University of Aeronautics And Astronautics$^1$, Tsinghua University$^2$ \\
\texttt{\small{jing.xu@nuaa.edu.cn, tbw23@mails.tsinghua.edu.cn, zhaohao@air.tsinghua.edu.cn}} \\
}

\iclrfinalcopy 
\begin{document}

\maketitle

\begin{abstract}

In this paper, we introduce a novel algorithm named Key Patch Proposer (KPP) designed to select key patches in an image without additional training. Our experiments showcase KPP's robust capacity to capture semantic information by both reconstruction and classification tasks. The efficacy of KPP suggests its potential application in active learning for semantic segmentation. Our source code is publicly available at: \url{https://github.com/CA-TT-AC/key-patch-proposer}.
\end{abstract}

\section{Introduction and Related work}

Masked Auto-encoder (MAE) \citep{he2022masked} has emerged as a potent strategy for representation learning. The strategy boosts the encoder's ability to produce non-trivial representations for images by dividing an image into non-overlapping patches, randomly masking the majority of the patches, and optimize the weights of the encoder so as to reconstruct the masked patches. MAE effectively primes the backbone network by learning textures within patches and interconnection between patches, improving the adaptability for a variety of downstream tasks.

One particular downstream task is active learning \citep{AL_best_practices}. In the history of active learning for 2D images, the acquisition of annotations is performed at various scale, e.g., the full image \citep{Sinha_Ebrahimi_Darrell_2019}, superpixels \citep{Superpixels}, polygons \citep{Mittal_Tatarchenko_Çiçek_Brox_2019} \citep{Golestaneh_Kitani_2019}, or even pixels \citep{pixels}.

Following the patch division in MAE, we consider the active learning at the patch scale for images and we propose a novel non-learning based patch proposal method, referred to as \textit{Key Patch Proposer} (KPP). The proposed patches, i.e., the key patches, hold significant potential for enhancing active learning in semantic segmentation tasks. The detailed task formulation is in Sec. \ref{section: method}.

Patch proposal fundamentally represents a submodular function maximization problem (SFMP) \citep{Submodular} \citep{Non-monotone}, which is NP-hard even in the absence of constraints. Details on the correlation between KPP and SFMP can be found in Appendix \ref{supp}. Deep learning methods are considered as solutions to this NP-hard problem, but the experiments suggest otherwise. In this paper, the proposed KPP is inspired by greedy search, a common polynomial-time approximation algorithm for SFMP, to obtain a sub-optimal solution for this inherently challenging issue. \cite{greedy_proof} has proved that greedy algorithm is one of the best solutions of SFMP in terms of performance level.

\section{Method}
\label{section: method}
Firstly, we resize an image into a fixed size ($224 \times 224$)  and then divide it into patches, following \cite{he2022masked}.
We define the problem of patch proposal as follows: given a set of patches $P$, we aim to find a subset of patches $P^* \subseteq P$ with $|P^*| = r |P|$ to minimize the L2 error $L$ of the image reconstructed using patch set $P^*$. Formally, we aim for a patch set $P^*$ that satisfies:
\begin{equation}
    P^* = \argmin_{\{P_s \subseteq P, |P_s| = r |P|\}} L(P_s)
\end{equation} 
where $L(P_s)$ represents the reconstruction error between the ground-truth masked patches $P  \setminus P_s$ and the masked patches predicted by a pretrained MAE with the unmasked patches $P_s$.
$r \in (0, 1) \subset \mathbb{R}$ is a hyper parameter that controls the ratio of proposed patches, and $|P|$ denotes the cardinality of $P$. 

We propose a greedy search method to solve this problem. The algorithm starts with an initial set $P_0$ which contains only the central patch. Then, the algorithm iteratively adds the patch that minimizes the reconstruction error. With $i$ being the iteration index and also number of patches in $P^*_i$,
\begin{equation}
    \begin{aligned}
        p^*_i & = \argmin_{p \in P \setminus P^*_{i-1}} L(P^*_{i-1} \cup \{p\}) \\
        P^*_i & = P^*_{i-1} \cup \{p^*_i\}
    \end{aligned}
\end{equation}
The algorithm terminates when the number of patches in $P^*_i$ reaches $r |P|$.
Pseudo codes are provided in Appendix \ref{supp_algo}.

\section{Experiments}
\begin{figure}[htb]
    \centering
    \includegraphics[width=\textwidth]{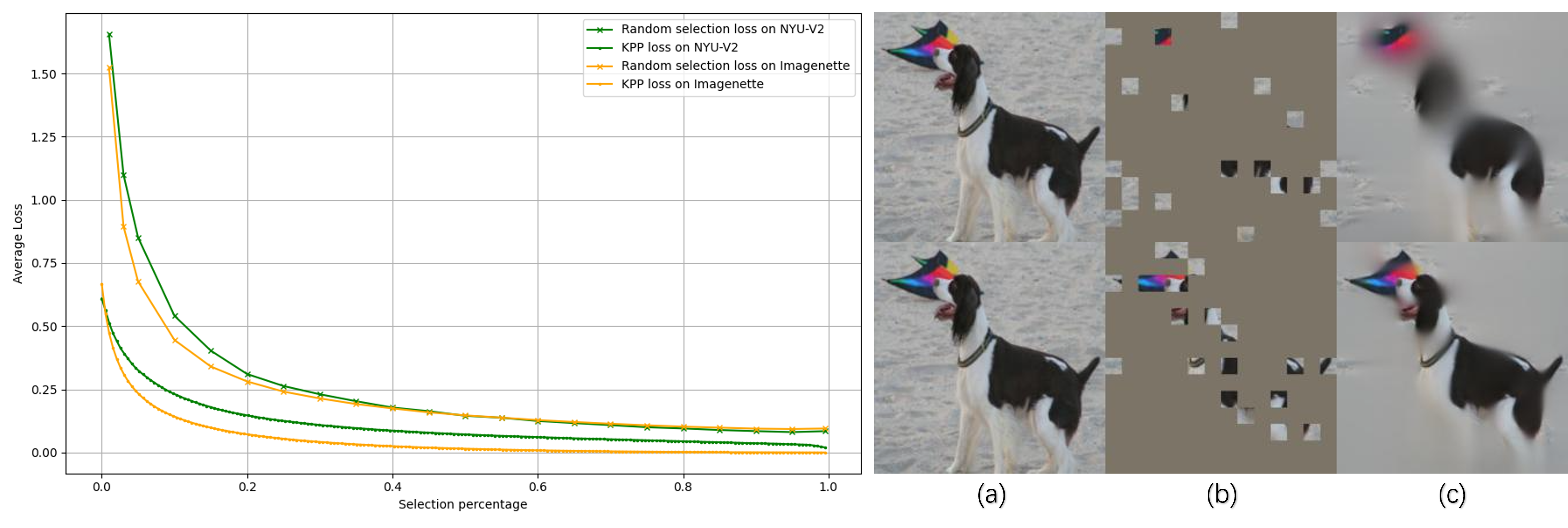}
    \caption{Left: Comparative analysis of reconstruction loss between KPP and random patch selection across various selection percentages. Right Section: The upper row demonstrates the random selection method, while the lower row depicts the KPP algorithm's approach. (a): ground truth; (b): selected patches; (c): reconstructed images utilizing the chosen patches.}
    \label{fig}
\end{figure}

We conduct experiments on ImageNette dataset \citep{imagenette}, and NYU Depth V2 dataset \citep{NYU} (See Appendix \ref{supply_dataset} for detail). We use the pretrained ViT-B/16 model provided by \cite{he2022masked} and the reconstruction loss is the mean squared error between the original masked patches and the reconstructed masked patches. We compare the performance of KPP with random patch selection. As demonstrated in the left chart of Figure \ref{fig}, we can see that KPP achieves lower reconstruction loss than random patch selection across various selection percentages. In the right section of Figure \ref{fig}, we visulize one case of random patch selection and KPP, with more shown in Appendix \ref{suppB}. Qualitative results demonstrate that KPP selects informative and representative patches of the original image.

\begin{table}[h]
    \centering
\begin{tabular}{l|l|l|l|l|l}
Selection percentage & 5\%   & 10\%  & 25\%  & 50\%  & 100\%                  \\ \hline
KPP acc              & 92.26 & 96.25 & 98.47 & 98.93 & \multirow{2}{*}{99.11} \\
Random acc           & 89.53 & 95.08 & 98.32 & 98.88 &                       
\end{tabular}
    \caption{Classification accuracy comparison of ViT-B/16 model on imagenette dataset.}
    \label{table1}
\end{table}

We also conduct quantitative experiments on ImageNette \citep{imagenette}. During both training and validation stages, we use KPP or random strategy to select input patches, and then finetune the pretrained ViT-B/16 on the selected patches only. Performances with and without KPP are reported in Table \ref{table1}, indicating that KPP surpasses random patch selection in terms of classification accuracy across various selection percentages, especially with small percentage of unmasked patches. 

\section{Conclusion}

In this work, we propose a novel algorithm, KPP, to select key patches within an image. Through our experiments focused on reconstruction and classification, we demonstrate the potential of KPP to identify key patches and quantify the semantic information they contain. We envisage KPP's application in active learning, a possibility we aim to further explore in future research.


\bibliography{iclr2023_conference_tinypaper}
\bibliographystyle{iclr2023_conference_tinypaper}
\newpage
\appendix
\section{Explaining on reduction of patch proposal problem to submodular function maximization problem}
\label{supp}
In mathematics, a submodular set function (also known as a submodular function) is a set function that, informally, describes the relationship between a set of inputs and an output, where adding another input has a decreasing additional benefit.

If $\Omega$ is a finite set, a submodular function is a set function $f : 2^{\Omega} \to \mathbb{R}$, where $2^{\Omega}$ denotes the power set of $\Omega$, which satisfies one of several conditions. One of the conditions is that for every $X, Y \subseteq \Omega$ with $X \subseteq Y$ and every $x \in \Omega \setminus Y$, we have that $f(X \cup \{x\}) - f(X) \geq f(Y \cup \{x\}) - f(Y)$. The purpose of this section is to prove that we can define a submodular function $f$ in the reconstruction task that satisfies the condition above.

In the reconstruction task, we define set $\Omega$ as all the patches in a image with positional information. $X$ is some selected patches from $\Omega$. We define $f$ as the mean squared error of the whole image but not of the masked patches. Formally, we define $f$ as follows:
\begin{equation}
    f(X) = (\text{MAE}(X) - Sort(\Omega))^2
\end{equation}
where $Sort(\Omega)$ is the sorted patches in $\Omega$ by their positional information. It can be easily verified that $f$ satisfies the condition above. Suppose $X \subseteq Y$ and $x \in \Omega \setminus Y$ is a new patch towards both $X$ and $Y$. $x$ always provides more semantics for the reconstruction of $Y$ than $X$, since $Y$ is a superset of $X$. Therefore, $f(X \cup \{x\}) - f(X) \geq f(Y \cup \{x\}) - f(Y)$.

By doing such verification, we can use polynomial-time approximation algorithms for submodular function maximization problem to solve the patch proposal problem.

\section{Algorithm Overview}
\label{supp_algo}
In this section, we provide pseudo codes to explain our proposed KPP algorithm briefly. $n_{keep}$ in line 1 represents the number of patches to be kept. We initialize $L_{min}$ with a large constant to ensure that it will be updated. In the for loop between line 5 and 10, we figure out the next best patch $p_i^*$ towards $P_{i-1}^*$ and then update $P_i^*$ accordingly.
\begin{algorithm}
	\caption{Pseudo-code of KPP algorithm.}
	
	\begin{algorithmic}[1]
		    \REQUIRE set $P$ with all patches in, proposal ratio $r$, empty set $P^*_1$ 
            \STATE $n_{keep} \leftarrow r|P|$
		\FOR{$i$ from $2$ to $n$}
		\STATE $L_{\min} \leftarrow \text{MAXN}$
                \FOR{$p$ in $P \setminus P_{i-1}^*$}
                    \STATE $L_{cur} \leftarrow \text{Loss}(P_{i-1}^* \cup \{p\})$
                    \IF{$L_{cur}<L_{min}$}
                    \STATE $p_i^* \leftarrow p$
                    \STATE $L_{min} \leftarrow L_{cur}$
                    \ENDIF
                \ENDFOR
	    \STATE $P_i^* \leftarrow P_{i-1}^* \cup {p_i^*}$
		\ENDFOR
  \RETURN $P_n^*$
	\end{algorithmic}
\end{algorithm}

\section{Dataset Overview}
\label{supply_dataset}
ImageNette dataset \citep{imagenette} is a subset of ImageNet \citep{imagenet} with 10 classes distributed in 9469/3925 training/validation images. 

 The NYU-Depth V2 dataset \citep{NYU} is comprised of video sequences from a variety of indoor scenes as recorded by both the RGB and Depth cameras from the Microsoft Kinect. It contains 1,449 RGB-D samples where
 795 image-depth pairs are used to train the RGB-D model, and the remaining 654 are utilized for
 testing. 

In the scope of our experiment, we exclusively utilize the RGB data from the NYU-Depth V2 dataset and restrict our experimental activities to its training set. This limitation is guided by the requirement in active learning systems for patch selection exclusively during the training phase. Our model, having been pretrained on ImageNet-1k, does not demonstrate zero-shot learning capabilities when deployed on ImageNette. Therefore, to validate its zero-shot learning potential, we employ the NYU-Depth V2 dataset. This approach substantiates the pretrained model's capability for direct application to other downstream datasets without necessitating further training.

\section{Ablation Study on Initial Patch}

\begin{figure}[htb]
    \centering
    \includegraphics[width=\textwidth]{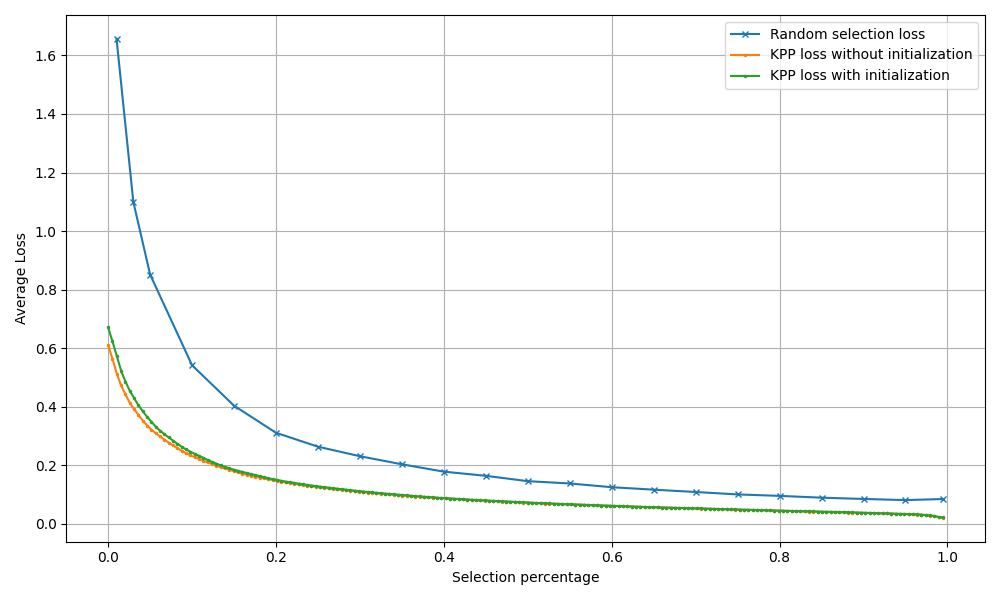}
    \caption{Ablation study of reconstruction loss with/without initial patch.}
    \label{ablation}
\end{figure}
We present an ablation study on the choice of the initial patch, depicted in Figure \ref{ablation}. This study was conducted using the training set of NYU-V2, akin to the scenario illustrated in the left chart of Figure \ref{fig}. Notably, our results demonstrate a clear distinction: the loss incurred without this initial patch selection is obviously lower compared to when this initialization step is included, particularly noticeable when the selection percentage is below 0.1.
\section{Qualitative Results for KPP and random selection}
\label{suppB}
\begin{figure}[htb]
    \centering
    \includegraphics[width=\textwidth]{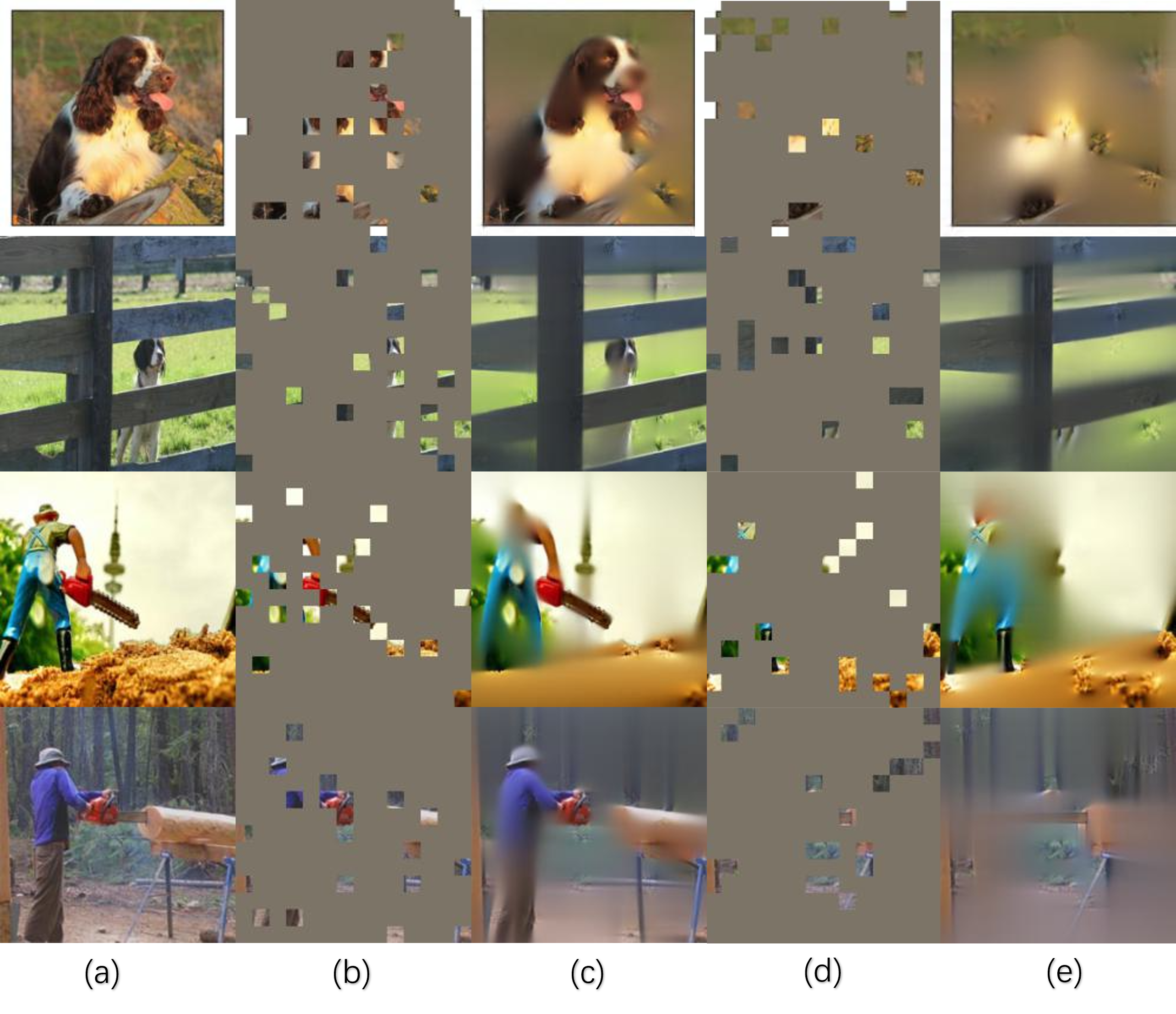}
    \caption{(a): Original images; (b): 10\% patches selected by KPP; (c): Images reconstructed by KPP patches. (d): 10\% patches selected randomly; (e): Images reconstructed by random selected patches.}
    \label{supply1}
\end{figure}
\end{document}